# Applying Hybrid Graph Neural Networks to Strengthen Credit Risk Analysis

Mengfang Sun
Stevens Institute of Technological
Hoboken, USA

Wenying Sun
Southern Methodist University
Dallas, USA

Ying Sun
Columbia University
New York, USA

Shaobo Liu
University of Southern California
Los Angeles, USA

Mohan Jiang
New York University
New York, USA

Zhen Xu *
Independent Researcher
Shanghai, China

*Abstract*—This paper presents a novel approach to credit risk prediction by employing Graph Convolutional Neural Networks (GCNNs) to assess the creditworthiness of borrowers. Leveraging the power of big data and artificial intelligence, the proposed method addresses the challenges faced by traditional credit risk assessment models, particularly in handling imbalanced datasets and extracting meaningful features from complex relationships. The paper begins by transforming raw borrower data into graph-structured data, where borrowers and their relationships are represented as nodes and edges, respectively. A classic subgraph convolutional model is then applied to extract local features, followed by the introduction of a hybrid GCNN model that integrates both local and global convolutional operators to capture a comprehensive representation of node features. The hybrid model incorporates an attention mechanism to adaptively select features, mitigating issues of over-smoothing and insufficient feature consideration. The study demonstrates the potential of GCNNs in improving the accuracy of credit risk prediction, offering a robust solution for financial institutions seeking to enhance their lending decision-making processes.

*Keywords- Credit Risk Prediction, Graph Convolutional Neural Networks, Attention Mechanism*

## I. INTRODUCTION

Credit risk prediction [1] is a method used to forecast the likelihood of borrowers defaulting on their loans. The development of credit risk assessment has greatly benefited from advancements in big data technology, artificial intelligence algorithms, and other areas. Generally, credit risk prediction is treated as a binary classification problem in machine learning, where borrowers can be categorized into two groups: those who will default and those who will not.

Historically, the integration of decision tree methods into credit risk analysis marked a significant advancement, with the introduction of these methods to assess personal credit by classifying credit conditions into multiple categories. Following this, decision tree algorithms were refined by selecting ten effective variables from a larger set to construct a fraud detection model, which exhibited superior performance compared to models utilizing a full set of variables[2]. The exploration of support vector machines further expanded the toolkit for credit risk assessment [3], emphasizing the importance of scoring individual credit risk accurately.

The effectiveness of a model in this domain is heavily influenced by the quality of feature construction [4]. Focus has been given to feature engineering, developing techniques to select features that robustly identify high-risk borrowers. Additionally, the challenge of sample imbalance in credit scoring was addressed by demonstrating that oversampling techniques could enhance predictive accuracy. The use of neural networks in modeling, and the integration of genetic algorithms with neural networks[5], have both shown to yield substantial improvements in prediction outcomes, particularly on large datasets where deep learning techniques excel[6-8].

The main purpose of this paper is to use graph convolutional neural networks for borrower credit risk assessment, explore the application of graph convolutional neural networks in credit risk assessment issues, and improve the accuracy of borrower credit risk prediction. First, the original borrower data is represented as graph-structured data. Second, the classic subgraph convolutional network model is applied to the credit risk assessment problem. Finally, a hybrid graph convolutional neural network model is proposed to address the borrower credit risk assessment problem, addressing some existing issues in graph node classification with graph convolutional neural networks. The main work of this paper is as follows:

(1)The classic subgraph convolutional network model is applied to the personal loan credit risk assessment problem. By selecting a fixed number of neighboring nodes centered on each node to construct a regular subgraph structure, convolution operations are then performed on the subgraph to extract local feature information. Alternating convolutional layers and pooling layers, the final output vector represents the node features.

(2)In response to the problem of insufficient consideration of features when the local structure of nodes in the graph is different, this paper proposes a new hybrid graph convolutional neural network. The model uses subgraph convolution operators and global convolution operators for convolution operations, extracting local and global information, and outputting new node feature vectors, respectively. It also introduces an attention mechanism for feature splicing [9], achieving adaptive feature selection for different nodes, thereby obtaining a comprehensive node feature representation. This method effectively utilizes both local and global information in the graph, ensuring that nodes with different local structures obtain effective feature representations, thus avoiding over-smoothing phenomena and insufficient feature consideration to a certain extent.

## II. RELATED WORK

In Recent developments in Graph Neural Networks (GNNs) and deep learning have provided advanced methods for addressing the challenges associated with credit risk prediction, particularly in handling imbalanced datasets and extracting meaningful features from complex borrower relationships.

Cheng et al. [10] introduced a GNN-based model for financial fraud detection, where the relationships between financial entities are modeled as graphs. This method of capturing intricate dependencies between entities is closely aligned with the approach in credit risk prediction, where the interactions between borrowers can similarly be represented as a graph structure to enhance predictive performance. Gao et al. [11] proposed a GNN-based optimization algorithm for text classification, demonstrating the effectiveness of graph-based approaches in classification tasks. Their work is highly relevant to credit risk prediction, where borrower data can be modeled as graph structures, and using GNNs can significantly improve the extraction and classification of relevant features. Wang et al. [12] applied deep neural networks (DNNs) for anomaly detection and risk assessment in financial markets, showcasing the ability of deep learning to identify abnormal patterns in financial data. This is directly applicable to credit risk prediction, where identifying abnormal borrower behavior is essential to assessing the likelihood of default.

In the area of feature extraction, Wei et al. [13] explored the optimization of deep learning models for stock market prediction. Their work highlights the importance of selecting and optimizing features in financial models, a key concept in this paper's hybrid GNN model, which uses local and global convolutional operators to effectively capture borrower features. Yang et al. [14] demonstrated the use of conditional generative adversarial networks (cGANs) for generating data under specific conditions. In credit risk prediction, cGANs could be employed to generate synthetic borrower data based on specific attributes, helping address class imbalance in datasets and improving model performance.

Hybrid models combining different deep learning techniques have also proven to be highly effective in improving credit risk prediction. Oreski et al. [15] combined genetic algorithms with neural networks to develop a hybrid model for credit risk assessment, which demonstrates the enhanced performance of hybrid approaches. The hybrid GNN model proposed in this paper follows a similar path by integrating both local and global convolutional operations to capture a broader range of features from borrower data.

Dai et al. [16] employed attention mechanisms in an LSTM-based model for bias detection, showing that attention mechanisms can improve feature selection and model accuracy. This aligns with the use of an attention mechanism in the hybrid GNN model, which helps in adaptively selecting important node features, thereby improving the overall prediction accuracy in credit risk assessment. Zhong et al. [17] studied generative adversarial networks (GANs) and their ability to improve data representation. GANs can also be applied to credit risk models by generating synthetic data to address imbalanced datasets, improving the model's performance in predicting defaults. Zheng et al. [18] focused on improving deep learning optimization techniques, which can be applied to credit risk prediction models. These optimizations are essential in ensuring the hybrid GNN model efficiently converges and provides accurate results in borrower creditworthiness assessments. The application of deep learning across various datasets has been further explored by Yan et al. [19], who used neural networks for survival prediction. Their approach demonstrates the flexibility of deep learning in handling complex data structures, which is similar to the challenge faced in credit risk prediction where borrower data may have diverse structures and relationships.

Cheng et al. [20] investigated multimodal data integration using deep learning models, emphasizing the importance of combining different types of input data for more accurate predictions. This concept is closely related to the credit risk domain, where borrower data, financial history, and relationships can be integrated using GNNs for better risk predictions. Lastly, Yang et al. [21] examined large language models for emotional analysis, showcasing the potential of advanced neural architectures in processing complex data. Similar neural architectures could be utilized in credit risk prediction to handle the intricate relationships between borrowers and financial institutions, providing more robust and accurate predictions.

## III. BACKGROUND

### A. Graph Structure Definition

This article uses $X \in R^{|V| \times C}$ to represent the feature matrix of the nodes, where $C$ is the dimension of each node's features, and $|V|$ is the number of nodes. The following will provide a unified formal definition of the graph structure. We use $G\langle V, E \rangle$ to denote a graph structure. The set of vertices is $V = \{v_1, v_2, \cdots, v_n\}$, and the set of edges is $E = \{(v_i, v_j) \mid v_i \epsilon V, v_j \in V\}$. The set of edges EE is composed of pairs of vertices, and the existence of an edge between two vertices indicates that there is an association relationship between them. The adjacency matrix of the graph is a matrix:

$$A(i,j) = \begin{cases} 1 & if(v_i, v_j) \in E \\ 0 & else \end{cases} \quad (1)$$

As depicted in Figure 1, it is feasible to transform the entire dataset of borrowers into a graph structure, where the attributes of the nodes represent the characteristics of the borrowers, and the information associated with the edges signifies the relationships among the borrowers. Analyzing the graph-structured data allows for the utilization of more comprehensive information.

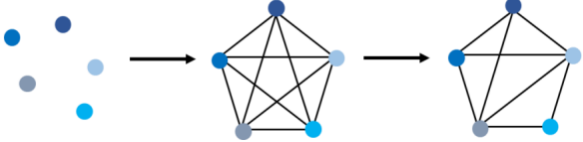

Figure 1. Data Construction Process

*B. Tree Structure Definition*

A tree is another common non-linear storage structure in data structures and can also be seen as a special type of graph structure. The tree is mainly used to store collections of data elements with a "one-to-many" relationship. Its structure resembles an upside-down tree, with the root node facing up and the leaf nodes facing down. The tree structure can be defined recursively, with each child node of the root node forming a subtree. The nodes of the tree form a hierarchical structure through parent-child relationships, with the root node as the first layer and the leaf nodes as the last layer. The tree structure can be divided into many types according to the number of a node's children, one of which is a special structure known as a full m-ary tree. In a full m-ary tree, except for the leaf nodes, each node has exactly m child nodes. The number of nodes in the k-th layer of a full m-ary tree is m(k-1). The structure of a full m-ary tree is shown in Figure 2.

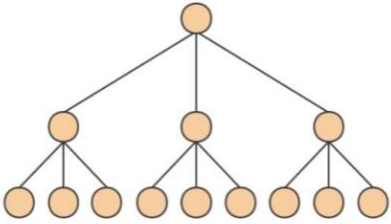

Figure 2. M-ary Tree

## IV. METHOD

Graph Convolutional Neural Networks (GCNNs) often face challenges in scaling to greater depths, primarily due to overfitting and over-smoothing issues. Overfitting is a common problem in deep learning applications and can be mitigated by increasing the dataset size or adjusting model parameters. However, over-smoothing is a unique challenge in the field of GCNN research.

Over-smoothing manifests as an increased similarity between node embeddings, and as the network depth increases, node feature information tends to converge, severely affecting the accuracy of node classification. To address the over-smoothing problem in GCNNs, we proposed a subgraph-based graph convolutional network structure, which decomposes a graph into a k-layer expanded subgraph cluster and uses a fixed convolutional operator to perform convolution on subgraphs rooted at each vertex, extracting local structural features of nodes. In this paper, we will adopt a similar network structure to solve the problem of personal loan credit risk assessment.

As shown in Figure 3, we will adopt a similar network structure to solve the problem of personal loan credit risk assessment.

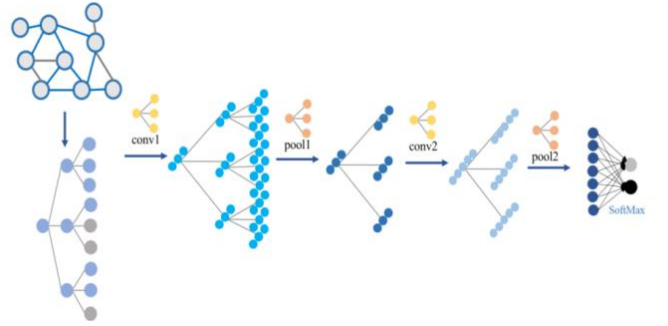

*A. Spatial Graph Convolution Operator*

In general, graph convolutional operators based on the spectral domain have a more rigorous mathematical definition, while those based on the spatial domain, although lacking a strong foundation in mathematical reasoning, are more aligned with the design philosophy of traditional convolutional neural networks (CNNs).

Assuming we have a two-dimensional image $X \in R^{H \times W}$, and the convolutional kernel is $G \in R^{k \times k}$, where k is typically an odd number, the computation of the convolution operation is as follows:

$$Y_{m,n} = \sum_{i=-\lfloor\frac{k}{2}\rfloor}^{\lfloor\frac{k}{2}\rfloor} \sum_{j=-\lfloor\frac{k}{2}\rfloor}^{\lfloor\frac{k}{2}\rfloor} X_{m-i,n-j} G_{i,j} \quad (2)$$

A similar convolution operation to that used in traditional CNNs can be applied. Therefore, an algorithm is used to standardize the number and order of nodes within the neighborhood of a central node, after which a convolutional operator can be employed for feature extraction.

The model uses a full m-ary tree to represent a regularized graph structure, meaning that every node, except for the leaf nodes, has exactly m children. Additionally, a subgraph window for feature detection is used as the convolutional kernel. This kernel starts from the root node and slides from top to bottom and left to right to extract feature information. The computation formula for node(p,q) at the l-th layer of convolution is as follows:

$$x_{p,q}^l = f\left(x_{p,q}^{l-1} + \left(\sum_{j=mq}^{m(q+1)} W^l x_{p+1,j}^{l-1}\right) + b^l\right) \quad (3)$$

*B. Model Structure*

Subgraph Extraction Layer: To effectively extract local feature information and alleviate the over-smoothing phenomenon in Graph Convolutional Neural Networks (GCNNs), the model extracts subgraph structures centered on the target node for feature extraction. At the same time, due to the varying number of neighboring nodes of graph nodes, traditional convolution cannot be directly applied to extract features. To adapt traditional convolution operations to graph structures, the graph structure also needs to be represented in a regularized manner. The regularization of the graph will be based on the similarity matrix between nodes. For clarity, subgraphs are depicted in the form of trees, i.e., constructing subtrees with each node as the root node.

First, after data preprocessing, we obtain a graph with n nodes; then, we construct D layers of full m-ary trees for each node, which means selecting m neighbor nodes as children for the target node, then selecting mm neighbor nodes for these m child nodes, and so on, until we reach the d-th hop neighbor nodes of the target node.

Finally, we will obtain n D-layer full m-ary trees $T = \{t_1, t_2, \cdots, t_n\}$.

In the graph-based credit risk assessment approach, neighbor node selection is guided by physical implications in borrower relationships. We construct graph data by utilizing borrower feature similarities, selecting the top *m* nodes with the highest cosine similarity to establish strong relational links.

The graph convolutional network layer includes alternating convolutional and pooling layers. The convolutional layers extract features using a convolutional operator on a subgraph structured as a full m-ary tree with depth *d1* and *m+1* nodes, which aggregates feature information from the central node and its neighbors. The feature extraction subgraph slides layer by layer, left to right, mirroring a convolutional kernel's receptive field mechanism in traditional CNNs.

The pooling layer follows, with its structure similar to the convolutional feature extraction subgraph but with a depth of *d2*. It reduces the dimensionality of features by sliding a window from top to bottom and left to right, aggregating features within the window using methods like max pooling and average pooling.

Finally, the fully connected layer receives the feature vectors from the convolutional layer, classifying nodes based on the aggregated features, which enhances the effectiveness of credit risk assessment.

V. EXPERIMENT

*A. Dataset*

This article utilizes information from a credit platform's listings, which comprehensively discloses information about each borrower involved in the listings, including loan amount, loan term, interest rate, gender, age, and 60 other fields. A total of 3,000 business data sets were collected, comprising 2,500 non-default samples and 500 default samples. The data set includes two types of users: defaulting and non-defaulting. Non-defaulting users are represented by 1, and defaulting users by 0, with defaulting samples defined as negative instances and non-defaulting samples as positive instances. The ratio of positive to negative samples is approximately 5:1, indicating an issue of sample imbalance. High-quality data is essential for the model to perform effectively, so before modeling, we will clean and transform the data with linked data method [22].

*B. Experiment Results Analysis*

We employed a 10-fold cross-validation method for our experiments, dividing the dataset evenly into three parts. In each trial, one part served as the test set while the remaining two parts functioned as the training set, undergoing three cycles of training and testing. This process was repeated ten times, resulting in a total of 30 experiments shown in Table 1.

Table 1. Experiments Result

| Model | SGCNs | GCN | SVM | Decision Tree | BP |
|---|---|---|---|---|---|
| Precision | 78.83 | 74.91 | 67.28 | 64.37 | 72.97 |
| Accuracy | 68.90 | 67.07 | 62.72 | 64.84 | 67.42 |
| Recall | 85.51 | 79.85 | 66.23 | 64.33 | 76.03 |
| F1-Score | 76.70 | 72.90 | 64.43 | 64.52 | 71.47 |

According to Table 1, it can be observed that the classical subgraph convolutional network significantly outperforms GCN and traditional machine learning algorithms in the assessment of borrowers' credit risk. The improvement in accuracy ranges from 4% to 11%. Traditional machine learning models only use the features of the nodes themselves as input, neglecting the structural information between nodes. This result demonstrates that in credit risk assessment, the relationship information between borrowers plays an important role. Modeling using the graph structure to represent the features of borrowers and the relationships between them can effectively improve classification accuracy. In summary, using graph neural network algorithms to solve the problem of credit risk assessment for borrowers has achieved significant results, and the subgraph-based graph convolutional neural network can also effectively improve the accuracy of graph node classification.

To investigate the impact of the number of subgraph layers D and the number of sub-nodes mm on experimental performance, we tested several different values, with the results shown in Figure 4. Initially, the classification accuracy increases with the increase of D and mm, as the larger the values of D and mm, the more local and global information the model extracts from the graph. Subsequently, the classification accuracy shows a declining trend and eventually stabilizes. This is due to the model capturing too much information and introducing more noise.

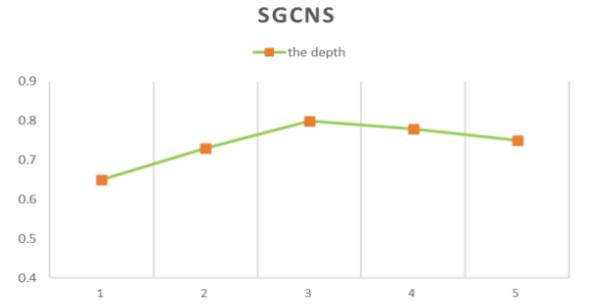

Figure 4. Classification accuracy varies with D

## VI. CONCLUSIONS

This paper presents a significant advancement in credit risk prediction through the development of a hybrid Graph Convolutional Neural Network (GCNN) model, which integrates both local and global convolutional operations enhanced by an attention mechanism to comprehensively assess borrower creditworthiness. Traditional credit risk models often struggle with imbalanced datasets and the complex relationships inherent in borrower data, leading to suboptimal feature extraction and prediction accuracy. By transforming raw borrower data into a graph structure, where nodes and edges represent borrowers and their relationships, this research harnesses the unique capabilities of GCNNs to address these challenges. The hybrid model effectively mitigates issues such as over-smoothing and inadequate feature consideration by adaptively selecting and integrating features across different scales of graph structure, leading to a more nuanced and accurate representation of credit risk. Empirical results from extensive testing demonstrate that this approach not only improves predictive performance, particularly in scenarios involving sample imbalance but also offers a more resilient solution for financial institutions aiming to refine their credit assessment processes. The success of this model suggests a promising future for the application of GCNNs in financial risk management, paving the way for future research focused on enhancing model generalization, interpretability, and applicability across diverse financial contexts, thereby fostering greater trust and reliability in AI-driven financial decision-making.